# Studying the Association of Online Brand Importance with Museum Visitors: an Application of the Semantic Brand Score

Fronzetti Colladon, A., Grippa, F., & Innarella, R.

# Studying the Association of Online Brand Importance with Museum Visitors: an Application of the Semantic Brand Score

*Fronzetti Colladon, A., Grippa, F., & Innarella, R.*

**Abstract**

This paper explores the association between brand importance and growth in museum visitors. We analyzed 10 years of online forum discussions and applied the Semantic Brand Score (SBS) to assess the brand importance of five European Museums. Our Naive Bayes and regression models indicate that variations in the combined dimensions of the SBS (prevalence, diversity and connectivity) are aligned with changes in museum visitors. Results suggest that, in order to attract more visitors, museum brand managers should focus on increasing the volume of online posting and the richness of information generated by users around the brand, rather than controlling for the posts' overall positivity or negativity.

**Keywords:** big data; social media; semantic brand score; brand importance; museum marketing.

1. **Introduction**

Competition in the tourism industry is increasingly based on high-volume data that is immediately available for travelers and tourism industry operators. In this increasingly knowledge intensive industry, big data represents an asset for various stakeholders, reducing information asymmetry for customers and increasing flexibility and responsiveness for organizations. Big data requires specific technology and analytical methods for its transformation into value (De Mauro et al., 2015). While recent studies used Google Trends Data to better understand tourist interests and intentions (Li, Pan, Law, & Huang, 2017; Padhi & Pati, 2017; Park, Lee, & Song, 2017), or focused on the analysis of online reviews (Fang, Ye, Kucukusta, & Law, 2016; Lee, Law, & Murphy, 2011; Wong & Qi, 2017), a less explored area is the analysis of the content extracted from online forums with the goal to predict museum visitors. Analyzing the content exchanged by users on sites such as TripAdvisor can help design promotional campaigns and brand awareness strategies that could inform and guide users' purchasing behaviors (Banerjee & Chua, 2016).

Leveraging the power of big data has the potential to reveal patterns and trends that are beneficial to several stakeholders in the tourism industry (Mandal, 2018; Xiang, 2018). Individual travelers can make decisions faster and use more complete and diversified information, which impacts the quality of their experience when choosing a destination to visit, comparing prices and building expectations for an upcoming trip (Leung, Law, van Hoof, & Buhalis, 2013).

Social media and online review sites support information search, decision-making and knowledge exchange for tourists and represent an opportunity for companies in the tourism industry to learn more about needs and find new ways to meet travelers' expectations (Gavilan, Avello, & Martinez-Navarro, 2018; Moro, Rita, & Coelho, 2017). Online travel forums are the ideal space for tourists to find answers to specific questions and link to resources to help them make the right decisions as they plan their travels (Hwang, Jani, & Jeong, 2013). Reviewers often share advice on practical matters, motivated by a desire for community empowerment, social support and joint-affirmation (Munar & Ooi, 2012).

The benefits of leveraging big data analytics to support strategic decision making in tourism destination management have emerged only over the past few years (Miah, Vu, Gammack, & McGrath, 2017), representing an interesting gap that this research would like to address. In this paper, we explore the potential application of a big data method to extract information from online travel forums. The goal is to evaluate the association between museum visitors and museum brand importance, by testing the potential value of new indicators that could be used to improve existing forecasting models.

We applied a measure defined Semantic Brand Score (SBS) that has been used to assess brand importance in other industries. SBS combines methods of semantic analysis and social network analysis to study large text corpora, across products, markets and languages (Fronzetti Colladon, 2018). Aligned with the work of Fronzetti Colladon (2018), we conceptualize online brand importance via three dimensions: prevalence, diversity and connectivity. These dimensions reflect respectively the frequency of use of a brand name (prevalence), such as a museum name, the heterogeneity of its textual brand associations (diversity) and its embeddedness at the core of a discourse (connectivity). To assess the benefits of adopting such a method to measure brand importance and relate it to museum visitors, in this paper we focus on five popular museums located in Italy, France and Hungary.

Consistently with this conceptualization of brand importance, our study aims to address the following research question: Is museum brand importance associated with variations of visitors?

The paper is organized as follows. Next section offers an overview of the theoretical background on social media, big data and text mining in the tourism industry. The third section describes the research design and methodology, providing a conceptual framework that highlights the hypotheses, and describing sample and data collection strategy. After illustrating our findings, the final section is devoted to discussing results and their implications.

## 2. Theoretical Background: Social Media, Brands and Tourism

Over the past ten years we have seen an increasing number of studies focused on the effect of social media on tourist decisions (Jacobsen & Munar, 2012; Miguéns, Baggio, & Costa, 2008), suggesting a positive linkage between perceived quality, electronic word of mouth, brand image, and brand performance (Barnes, Mattsson, & Sørensen, 2014). The high-context interactions offered by social media platforms impact consumers' attitudes and behaviors, leading them to perceive the brand in more positive terms, which ultimately increases purchase intentions (Kim & Ko, 2012). By interacting with others on a forum, users can reduce misunderstanding on what a brand offers, they can change their attitude towards a brand from negative or neutral to positive, as well as receive additional details that will lead them to finalize the purchase. Recent studies focused on specific aspects of brand equity, such as brand popularity (Gloor, 2017; Gloor, Krauss, Nann, Fischbach, & Schoder, 2009). Others applied methodologies to web data and focused on social interaction via social media (De Vries, Gensler, & Leeflang, 2012), or studied links among webpages or user generated content (Yun & Gloor, 2015).

Brand-based social media activities generates online or electronic word of mouth that improves the understanding of products and services thanks to the exchange of ideas among users, which improves marketing productivity and performance (Filieri & McLeay, 2014; Keller, 1993; Torres, Singh, & Robertson-Ring, 2015). A recent study on the impact of user interactions in social media (Hutter, Hautz, Dennhardt, & Füller, 2013) found that users' engagement with a Facebook page positively impact their brand awareness, word-of-mouth engagement and even purchase intentions. In a study focused on user-generated content and the effects of electronic word-of-mouth to hotel online bookings, Ye, Law, Gu and Chen (2011) found that online reviews have a significant impact on online sales, with a 10 percent increase in traveler review ratings boosting online bookings by more than five percent. Yang, Pan, Evans and Lv (2015) used web search query volume on Google and Baidu to predict visitor numbers for a popular tourist destination in China and demonstrated the predictive power of using search engines in understanding the travel process of tourists. Other studies explored the impact of good vs. bad ratings during the first stage of the decision-making process when travelers book a hotel, and found that web users tend to select hotels that have better ratings (Gavilan et al., 2018; Sparks & Browning, 2011). Similarly, Neirotti, Raguseo and Paolucci (2016) demonstrated how recommendations posted on social media by peers can positively influence travelers in choosing hotels and destinations that are consistent with their preferences and attitudes. These approaches, however, have some limitations when applied to contexts containing text that is not associated to a social interaction (for example a press release). We found a limited number of studies that linked semantic analysis with factors impacting brand equity.

**2.1. Text Mining in Tourism**

Assessing brand equity and importance has been usually done via market surveys administered to consumers and other stakeholders, or via financial methods (Belén del Río, Vázquez, & Iglesias, 2001; Lassar, Mittal, & Sharma, 1995). Surveys and traditional financial methods have limitations due to perception biases, sampling methods and excessive dependence on historical variables. An increasing number of studies are adopting text mining techniques and sentiment analysis approaches to study the informative contribution of travelers and users in online forums, with only few of them focused on museum visitors (Volcheck, Song, Law, & Buhalis, 2018). Aggarwal, Vaidyanathan and Venkatesh (2009) extracted information using the Google search engine and lexical text analysis to explore online brand representations, by examining the association between brands and a selection of adjectives or descriptors. Others applied text analytics to online customer reviews collected from Expedia.com to understand hotel guest experience and its relationships with guest satisfaction (Xiang, Schwartz, Gerdes, & Uysal, 2015). Their results show that the association between satisfaction rating and guest experience is strong, and that a general pattern can be observed between customers' use of particular words to describe the experience and the quality of service provided. Another interesting example of text mining applied to understand tourism demand is offered by the application of content analysis to 220 samples of Lonely Planet postings to assess the messages' functional information (Hwang et al., 2013).

Researchers have analyzed users' interactions by mining forum posts, mailing list archives, hyperlink structure of homepages, or co-occurrence of text in documents, though fewer have explored how content analysis on social media could be used in an integrated way to understand brand importance online. Text mining of online tourism reviews offers invaluable - and otherwise difficult to collect - review evaluations supporting comparative analysis (Bucur, 2015; Hu & Liu, 2004).

All the empirical studies on travelers' online behavior and its impact on economic performance that we have presented thus far have been heterogeneous and focused on a multiplicity of big data approaches. Most of the text mining systems and approaches developed in the past few years are based on an extraction of reviews from page content, and then use algorithms or text mining modules to process the content through a classification of reviews as positive, negative and neutral (Capodieci, Elia, Grippa, & Mainetti, 2019; Zhang, Fuehres, & Gloor, 2011). The framework that we use in this study (Semantic Brand Score ) goes beyond the textual classification of words or comments on social media, and incorporates new metrics of text analysis with indicators developed in the fields of social network analysis and semantic analysis (Fronzetti Colladon, 2018).

### 2.2. An integrated framework to study brand importance

The Semantic Brand Score (SBS) is a comprehensive framework based on widely accepted brand equity models (Keller, 1993; Wood, 2000) that evaluates brand importance using a composite approach that goes beyond counting the number of likes to Facebook brand pages or the number and valence of comments on social media. We will use the SBS composite indicator as the basis for our conceptual framework. The SBS is calculated based on three dimensions: prevalence, diversity and connectivity. Partially connected to the dimension of brand awareness (Aaker, 1996), prevalence represents the frequency with which the brand name appears in a set of text documents: the more frequently a brand is mentioned, the higher its prevalence. Prevalence used the frequency by which a museum name is mentioned in a discourse, and can be intended as a possible proxy for brand awareness (Keller, 1993). Keller's definition of brand equity and brand awareness includes the concept of differential response to knowledge of a brand name, suggesting that brand awareness is the starting point to building a positive image (Keller, 1993). The second SBS dimension, diversity, is related to the concepts of lexical diversity (McCarthy & Jarvis, 2010) and word co-occurrences (Evert, 2005). It measures the heterogeneity of the words co-occurring with a brand, assigning higher diversity to brands embedded in a richer discourse. *'A brand could be mentioned frequently in a discourse, thus having a high prevalence, but always used in conjunction with the same words, being limited to a very specific context'* (Fronzetti Colladon, 2018, p. 152). The more network neighbors a brand has, the more heterogeneous is the semantic context in which the brand name is used. This measure is higher when brand associations are more diverse and is consistent with previous research showing the positive effect of a higher number of associations on brand strength (Grohs, Raies, Koll, & Mühlbacher, 2016). The third component, connectivity, expresses how often a word (in our case a brand) serves as an indirect link between all the other pairs of words, while constructing a co-occurrence network (see Section 3). It reflects the embeddedness of a museum name in a discourse and can be considered as the expression of the connective power of a brand name, i.e. the ability to indirectly link different words and/or topics. This dimension is consistent with other studies. For example, Gloor et al. (2009) mapped semantic networks extracted from the web and found that the betweenness centrality of a brand could be used as a proxy for its popularity. Similarly, another study found that brand relevance in specific contexts can be measured via its betweenness centrality (Fronzetti Colladon & Scettri, 2019). While a brand name could be frequently mentioned (high prevalence) and might have heterogeneous associations to other brands or concepts (high diversity), the museum name could still be peripheral and not connected to the core of the conversations.

Overall, these considerations suggest that a museum, whose brand is frequently used in online forums (prevalence), that is embedded in a rich discourse (diverse), and is at the core of a discourse (connected), has a greater competitive advantage over other museums with a lower brand importance in terms of the ability to attract customers to their site. When users perceive that a brand is supported and mentioned by other users, who offer detailed explanations of the brand value, or add a comparative analysis, they are more likely to be persuaded and they might follow a similar path. We therefore formulate the first hypothesis as follows:

*H1: There is a positive association between the importance of museum brands in online conversations, measured through the Semantic Brand Score, and the number of museum visitors.*

An important brand is at the core of a conversation, with the possibility of being associated to either negative or positive feelings. On the other side, a brand that is used marginally, or that is very peripheral in a set of documents, can be classified as unimportant. When positive words are used to talk about a brand online (i.e. positive sentiment), word of mouth will likely lead to an increase of visitors for the museum. Therefore, a positive sentiment of the words which co-occur with a museum name should reinforce and complement its brand importance.

As suggested by different authors (Dellarocas, Awad, & Zhang, 2004; Smith, Menon, & Sivakumar, 2005), online reviews can be perceived as more credible than traditional word-of-mouth as they are originated by users with similar attitudes and preferences. The more users talk about a brand in positive terms, the higher the chance for the online word of mouth to be translated in economic value for the brand/museum. In their process of searching for information and validation and select their final tourism destination, users might change their initial choice based on the positive words associated with a brand that they find on the forum. Secondly, when users leave a positive review or a positive rating for a product, which seems to happen more often than leaving a negative comment (Bilro, Loureiro, & Guerreiro, 2019), other users will be more likely to buy that product or like that brand (Hollebeek, Glynn, & Brodie, 2014).

These considerations would lead us to assume that an increase in the positive sentiment associated with museum brands benefit the institutions by bringing in more visitors attracted by the peer-review validation. We therefore propose the following hypothesis:

*H2: The more positive the sentiment of online posts about a museum, the higher the number of visitors.*

Figure 1 illustrates the conceptual framework and hypotheses driving this study:

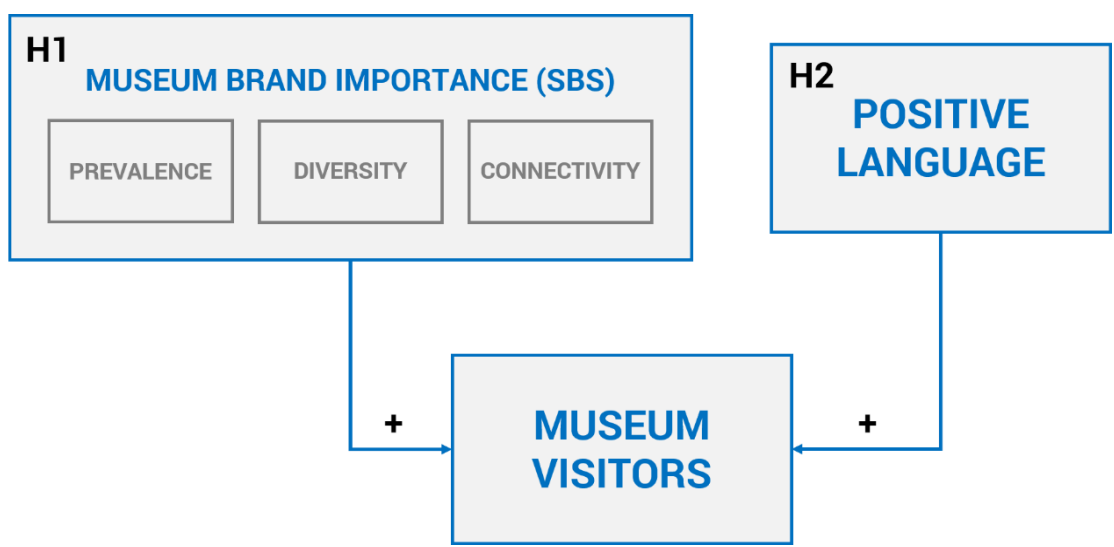

**Figure 1**. Conceptual Framework

3. Research Design

3.1. Data Collection

In this study, we considered the last 10 years of the online conversations happening on popular online forums, accessible also to non-registered users, where people exchange travel tips and opinions and share personal experiences on places and attractions. Travelers search online to increase the quality of future trips and to minimize potential risks associated with future travel (Jacobsen & Munar, 2012). A recent Nielsen report on social media (Shannon, Andrew, & Maeve, 2016) claim that travel websites were the second most-trusted source of brand information after recommendations from friends and family members.

For our analysis, we used an existing database comprising a large number of forum posts (Innarella, 2018). This database contains data on a selection of European museums: the Louvre and the Pompidou Centre in Paris, the Borghese Gallery and the Vittoriano in Rome, and the National Gallery in Prague. The data is made of more than 2,830,000 forum posts about tourist attractions in Paris, Rome and Prague, written by more than 113,700 users over ten years, from January 2007 to December 2016. Based on this existing database, we built a uniform naming conventions given the use of abbreviations for museums and attractions, and we corrected typing errors on some of the posts. The author's choice of selecting museums from the capital cities of France, Italy and Hungary was mostly experimental and explorative, with the aim of studying cities with a different attractive power, different attractions and characteristics. As reported by the United Nations World Tourism Organization, France and Italy were respectively first and fifth in terms of international arrivals, while Hungary increased its yearly international arrivals by 7% (*UNWTO Tourism Highlights: 2017 Edition*, 2017). A brief description of the museums included in the sample and the number of posts collected is presented in Table 1.

Information about museum visitors was collected considering the annual reports published on the website of each museum, as well as consulting the cultural aggregators Statistica Beniculturali and Egmus (http://www.statistica.beniculturali.it; http://www.egmus.eu). Since data was available on annual bases, our dependent variable consists of 50 observations (5 museums x 10 years). To build data consistency, we calculated our predictors, sentiment and the Semantic Brand Score, with an annual frequency (e.g. the brand importance of Louvre in 2016). The database we accessed only included posts written in English. Limiting the analysis to one single language was important in our study, in order to be consistent in the measurement of semantic variables. Although the calculation of both sentiment and Semantic Brand Score can be adapted to multiple languages, it would be inappropriate to directly compare scores coming from posts written in different languages. In future replications of this study, we aim to see whether standardization would be sufficient to address this issue, or whether alternative approaches are feasible. Moreover, English was the most frequently used language by tourists of different nationalities (Innarella, 2018). In order to appropriately measure brand importance, we analyzed all posts about Paris, Rome and Prague, even those not including museum names. This is particularly relevant for measures such as connectivity, which require to assess the position of a museum name in the co-occurrence network with respect to the general discourse.

| MUSEUMS | DESCRIPTION |
|---|---|
| **LOUVRE, PARIS** | The world's largest art museum and a historic monument in Paris, France. In 2017, the Louvre was the world's most visited art museum, receiving 8.1 million visitors. |
| **BORGHESE GALLERY, ROME** | Art gallery in Rome, Italy, housed in the former *Villa Borghese Pinciana*. It houses a large part of the Borghese collection of paintings, sculpture and antiquities; it has twenty in rooms across two floors. |
| **NATIONAL GALLERY, PRAGUE** | The most important gallery in the Czech Republic with the largest collection of Czech and international art. The collections are presented in a number of historic structures within the city of Prague, as well as other places. |
| **VITTORIANO, ROME** | The *Altare della Patria* (Altar of the Fatherland), also known as *Il Vittoriano*, is a white marble monument located in Rome, Italy, built in honor of Victor Emmanuel, the first king of a unified Italy. |
| **POMPIDOU CENTRE, PARIS** | A complex building designed by Renzo Piano and Richard Rogers, is home to the National Museum of Modern Art in Paris and is internationally renowned for its 20th and 21st century art collections. |

**Table 1**. Museums included in the sample.

### 3.2. Study Variables

We pre-processed text data in order to remove stop-words (i.e. those words which usually provide little contribution to the meaning of a sentence, such as the word 'and'), punctuation and special characters. We changed every word to lowercase and extracted stems by removing word affixes (Jivani, 2011), by using the NLTK Snowball Stemmer algorithm (Perkins, 2014). To conduct these preliminary operations and to calculate the SBS indicator, we adopted the programming language Python. The most important libraries we used for network analysis task were NLTK (Perkins, 2014), for Natural Language Processing, and Graph-Tool (Peixoto, 2014).

The subsequent step was to transform text documents into a social network where nodes are words that appear in the text. An arc exists between a pair of nodes if their corresponding words co-occurred at least once; arc weights are determined by the frequency of co-occurrence. Following this procedure, we obtained 30 networks: 3 city forums (Paris, Rome and Prague), 10 years of conversations. In order to filter out negligible or less frequent co-occurrences, we retained only the arcs which had a minimum weight of 5. Based on methods used by previous studies (Fronzetti Colladon, 2018), we adopted a five-word window for the determination of co-occurrences maximum range.

Prevalence was calculated as the frequency by which a museum name was mentioned in the forum posts. Diversity is a measure of the heterogeneity of textual brand associations and is higher when brand associations are more diverse. Diversity has been operationalized through the degree centrality measure (Wasserman & Faust, 1994):

$$Diversity(museum_i) = d(g_i)$$

It corresponds to the degree of the node $g_i$ which represents the museum brand: $d(g_i)$.

Connectivity reflects the 'brokerage power' of a museum name in the discourse about city attractions (Fronzetti Colladon, 2018). While a brand name could be frequently mentioned (high prevalence) and might have heterogeneous associations to other words (high diversity), the museum name could still be peripheral in the conversations. Connectivity, calculated as the betweenness centrality of the brand term (Fronzetti Colladon, 2018; Wasserman & Faust, 1994), can be considered as the expression of its connective power, i.e. the ability to indirectly link different words or groups of words (sometimes seen as discourse topics):

$$Connectivity\ (museum_i) = \sum_{j<k} \frac{d_{jk}(g_i)}{d_{jk}}$$

with $d_{jk}$ equal to the number of the shortest paths linking the generic pair of nodes $g_j$ and $g_k$, and $d_{jk}(g_i)$ equal to the number of those paths which contain the museum brand node $g_i$. As suggested in a more recent work of Fronzetti Colladon (2019), in the computation of connectivity we considered the inverse of arc weights in the determination of shortest network paths, and therefore calculated weighted betweenness centrality, using the algorithm proposed by Brandes (2001).

To compare measures derived from different networks (i.e. one for each year and for each museum), we standardized the values of prevalence, diversity and connectivity. Standardization was carried out subtracting the median to each individual score (of words in each network) and dividing it by the interquartile range. The Semantic Brand Score was subsequently calculated as the sum of the standardized values of its components (Fronzetti Colladon, 2018). According to this standardization procedure, SBS scores can either be positive or negative – based on the importance of a certain term, i.e. a museum name. If a term had a negative score, it means that its unstandardized value is below the median of the scores obtained by the other significant words in the discourse.

Each of the above mentioned measures was calculated as the variation with respect to the previous year. A first differencing of the variables permitted the elimination of time trends and produced stationary data. A first differencing was also applied to the dependent variable of our study, i.e. the yearly number of museum visitors.

Lastly, we measured the sentiment of museum brands, considering the valence of their textual associations, obtained from the polarity scores included in the SenticNet 4 dictionary (Cambria, Poria, Bajpai, & Schuller, 2016). As each association had its own strength – represented by the co-occurrence frequency – overall brand sentiment was calculated as the weighted average of association polarity. This measure has been subsequently rescaled in the range [0,1] with values below 0.5 representing a negative sentiment on average, and values above this threshold indicating a prevalence of positive associations. Other approaches for the calculation of sentiment are also possible, and could be tested in future research. For example, one could train an ad-hoc classifier, using supervised machine learning algorithms. Another alternative is to use the VADER lexicon included in the NLTK package, which seems to work well for texts extracted from social media (Hutto & Gilbert, 2014). We additionally tested this approach, without drawing significantly different conclusions from our models.

Table 2 presents the descriptive statistics for our variables. A first interesting result is that the average sentiment of the textual brand associations was positive, indicating that museums were mentioned on the forums with a usually positive valence of the words used. This might indicate that users have on average a positive attitude when they ask for or provide general information about museums (e.g. ticket price, location, opening hours, events and temporary exhibitions), with few posts expressing criticism. Standardized values of the SBS and its components are positive on average, suggesting that museum names got significant attention within the discourse about the three capital cities we analyzed.

| **Variable** | **M** | **SD** | **Min** | **Max** |
|---|---|---|---|---|
| Visitors | 2568946 | 3249613 | 18215 | 9437744 |
| Prevalence | 184.156 | 274.296 | 0 | 830 |
| Diversity | 7.276 | 9.083 | -.163 | 27.109 |
| Connectivity | 122.166 | 216.580 | -.002 | 757.846 |
| SBS | 313.598 | 494.676 | 0 | 1594.787 |
| Sentiment | .579 | .083 | .411 | .917 |

**Table 2.** Descriptive Statistics

## 4. Results

Our findings indicate that, as the SBS grows, so does the number of museum visitors, whereas a decrease in SBS is associated with a decrease of visitors. Figure 2 represents a contingency table, which shows the concordance of yearly change of SBS with change in the number of museum visitors. Blue squares indicate concordant cases, whereas grey square discordant ones.

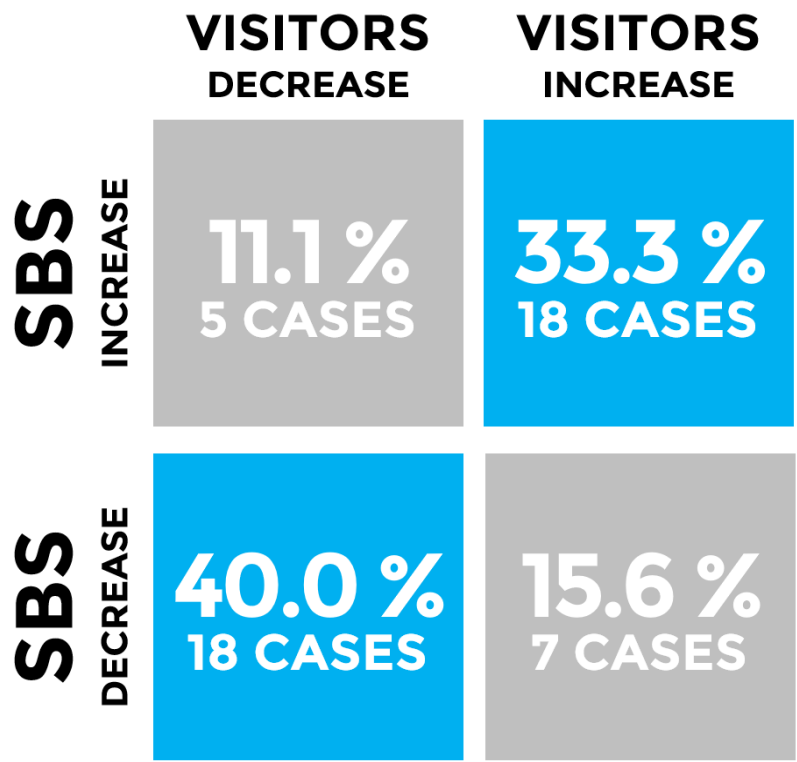

**Figure 2.** Association of Visitors and SBS Variations

In about 73% percent of cases the signs of these variations are concordant, supporting the idea that an increase in the SBS can be indicative of a higher number of museum visitors, whereas a decrease in SBS often associates with a decrease of visitors. These results are also confirmed by the significant Pearson Chi-square test ($\chi^2 = 9.82$, p = .002) and by the Fisher's exact test (p = .003), both carried out on the contingency table of Figure 2.

We subsequently built the multiple regression models presented in Table 3 to see change in which variables could better explain change in museum visitors. In all the models, we controlled for the possible effect of time and included several dummy variables representing each museum considered in our study. This method was appropriate since the selected museums had a different average number of visitors per year and we had repeated measures over time. Another possibility to deal with repeated measures over time would be to use multilevel regression models (Hoffman & Rovine, 2007; Singer & Willett, 2003), as these models consent an analysis of variance on multiple levels, to see which part is accountable to the differences in museums and which is residual. However, we built such models with respect to our dependent variable – nesting repeated measures (level 1) in museums (level 2) – and found that the intra-class correlation coefficient was close to zero, indicating a high dominance of the residual variance. In Model 1, we only included the control variables which we included in all models. In Models 2-5, we tested separately the three components of the SBS as well as the sentiment variable. Model 6 is the full model. In Models 7, we used the SBS instead of its separate components, and then combined it with sentiment in Model 8. Model 1 shows that control variables alone can explain about 17% of variance, however with a very low adjusted R-squared (.05) and only Time is significant. In the subsequent models, we tested separately the contribution of each dimension of the SBS, finding that prevalence and connectivity are the most significant, together with the SBS aggregate measure, all presenting a positive coefficient. Prevalence seems to be the most important predictor, being able to improve the variance explained by Model 1 of about 26%, and its adjusted R-squared of about 33%. Combining the three dimensions in Model 6 led to the best results, with an R-squared of 57.55% and an adjusted R-squared of 45.23%. In this model, diversity becomes significant. We checked for multicollinearity problems and found no evidence to support them (maximum Variance Inflation Factor was 2.33 for Model 6). On the other hand, as the SBS is the sum of standardized prevalence, diversity and connectivity, collinearity problems would arise if putting in the same model the three dimensions together with the final indicator. Following the suggestion of Fronzetti Colladon (2018), we additionally explored the impact of the sentiment variable, which resulted always non-significant.

|  | Model 1 | Model 2 | Model 3 | Model 4 | Model 5 | Model 6 | Model 7 | Model 8 |
|---|---|---|---|---|---|---|---|---|
| Time |  | -41679.2* | -12994.2 | -42565.3* | -32454.58 | -41595.91* | -15424.16 | -25979.1 | -25890.43 |
| Louvre | -135870 | -36534.6 | -139122.2 | -143816.2 | -135415 | -65345.1 | -120660.6 | -120183.7 |
| National Gallery Prague | 113552.8 | 56834.2 | 114846 | 101698.2 | 106752.9 | 55126.49 | 88847.29 | 81754.48 |
| Complesso del Vittoriano Rome | -3297.6 | -17052.95 | -2632.497 | -2244.497 | -89.539 | -8716.255 | 5480.918 | -2136.305 |
| Centre Pompidou | 78952.3 | 88850.9 | 77718.75 | 76387.36 | 77513.08 | 68995.43 | 78856.59 | 77355.93 |
| Prevalence |  | 5129.8*** |  |  |  | 6758.791*** |  |  |
| Diversity |  |  | 8115.49 |  |  | 142279.4** |  |  |
| Connectivity |  |  |  | 1185.371* |  | 511.5725 |  |  |
| SBS |  |  |  |  |  |  | 1164.427** | 1164.563** |
| Sentiment |  |  |  |  | 87153.26 | 16047.88 |  | 90871.11 |
| Constant | 254434.8 | 96251.36 | 259252.1 | 198028.2 | 203777 | 96802.04 | 162426.9 | 109597.4 |
| R-Squared | 0.1714 | 0.4317 | 0.1721 | 0.2853 | 0.1719 | 0.5755 | 0.3423 | 0.3428 |
| Adj R-Squared | 0.053 | 0.3314 | 0.026 | 0.1592 | 0.0257 | 0.4523 | 0.2262 | 0.2034 |

*Note*. *p < .05; ** p < .01; *** p < .001.

**Table 3.** Regression Models

We additionally checked for the effects produced by time lags of our variables, always considering their first differencing. However, these did not lead to better models, also considering the limited number of observations in our sample (50 total, 10 per each of the 5 museums). This finding is consistent with previous work showing that future visitors usually consult the latest forum posts about the topic of their interest while older reviews are perceived as less informative (Wu, Che, Chan, & Lu, 2015); therefore, in most cases, posts written in the same year of their visit, not before. Older posts also rarely appear on the first page of a forum search query.

Lastly, since the National Gallery of Prague is probably less known than the other museums, we tried to remove it from the analysis, to check the robustness of our models. The new results were fully consistent with those of Table 3, with a slight improvement of the Adjusted $R^2$ (from 0.4523 to 0.4659, for the best model, Model 6).

Multiple regression was a first attempt to prove the significance and directionality of the association of brand importance with museum visitors. Using a Naïve Bayes algorithm (John & Langley, 1995), we extended the analysis, obtaining predictions of change in museum visitors which have a reasonable accuracy. We used the machine learning software Weka (Holmes, Donkin, & Witten, 1994) and – after testing many combinations of algorithms, including Random Forests and Support Vector Machine (Breiman, 2001; Suthaharan, 2016) – we found that the best results were those of Naive Bayes (John & Langley, 1995). This produced forecasts of positive/negative change in visitors which were 75.56% accurate and had the positive or negative variations of the SBS and its components as predictors. Our choice was also supported by the results obtained using the Auto-Weka package (Kotthoff, Thornton, Hoos, Hutter, & Leyton-Brown, 2017). A reasonable fit of the Naive Bayes algorithm was confirmed by the average values of the Cohen's Kappa and of the area under the ROC curve – respectively 0.51 and 0.75. The algorithm was trained on a random 70% of the sample, and the remaining 30% was left out for evaluation. The process was repeated 500 times and the accuracy values we reported are on average.

**5. Discussion and Conclusions**

The recent availability of real-time, high-volume data, and the easy access to users' digital traces, offers a new opportunity to obtain improved understanding of tourist behaviors. This paper has described the application of a methodology based on the integration of social network analysis and text mining to measure brand importance and study its association with museum visitors: to this purpose we used the Semantic Brand Score (Fronzetti Colladon, 2018).

Our models support the first hypothesis and offer insights on the association between brand importance and change in museum visitors. The strongest proportion of variance explained in the regression models (57.6%) was obtained by combining the three dimensions of the SBS. Connectivity partially contributed to the increase of variance explained, but was not significant in the final regression model. However, connectivity, together with the other SBS dimensions, was important to improve the accuracy of the Naïve Bayes algorithm. The sentiment indicator, on the other hand, was never significant. It seems that when users provide articulated reviews on a museum – frequently mentioning its name, likely adding detailed explanations of pros and cons, and using heterogeneous words – the likelihood to convince others to finalize a purchase or reserve a museum pass is higher. This is aligned with a recent study on online consumer perception, review factuality and source credibility (Filieri, Hofacker, & Alguezaui, 2018), which suggests that consumers tend to look for reviews that report accounts of facts and events related to their experience. A study focused on Yelp.com comments provided similar evidences, showing that cognitive processing is more relevant than other components of brand affection and activation/energy (Bilro et al., 2019). Future tourists might be persuaded to follow other users' leads on a brand when information provided is more articulated, diverse and frequently discussed in several posts.

Our models suggest that what really matters in terms of building a strong and attractive brand is that users talk about the brand on social media, even if they provide comments that are not necessarily the most positive. Therefore, our second hypothesis was not supported by the models, since the sentiment of the words associated with a museum brand is not necessarily associated with an increase in its visitors over time. This contrasts with other studies indicating that purchasing behavior could be influenced by positive comments left by others (Kim & Ko, 2012). It seems that consumers tend to prefer posts that display rich information, rather than overly positive reviews, which is a result confirmed by other studies (Filieri et al., 2018; Filieri & McLeay, 2014).

The evidences collected in this study suggest that in order to increase museum visitors over time it is important to increase the volume of online posting and the richness of information generated by users around the brand. This seems to suggest that tourist might be influenced by the *awareness effect* generated by online word-of-mouth: the presence of brand names. In order to increase museum visitors over time, it is important to increase the volume of online posting, rather than controlling for the positivity or negativity of the posts. This is aligned with findings by Duan, Gu and Whinston (2008) who found a positive association between online word-of-mouth and movie sales: whereas box office sales were significantly influenced by the volume of online posting, higher ratings did not lead to higher sales. The simple presence of brand reviews conveys the existence of the product which makes it more desirable by consumers (Godes & Mayzlin, 2004; Viglia, Minazzi, & Buhalis, 2016). Since the sentiment of the words associated to the brand was on average positive, we can imply that only a few users had expressed negative comments on the museums. This could mean that a positive sentiment on average is sufficient to attract new visitors and that being even more positive is not necessary to promote a positive brand image. It would be interesting to replicate this study in scenarios where sentiment variations are more pronounced.

Museums today perform their functions in an extremely competitive market environment, with some of them struggling to survive due to decreasing visitor numbers and financial bottlenecks in the public sector (Gretzel, Werthner, Koo, & Lamsfus, 2015; Kovaleva, Epstein, & Parik, 2018). Implementing management techniques typically adopted in the for profit sector and designing new brand management strategies has the potential to increase the likelihood of repeat visits and recommendations to visit through word of mouth. For the past two decades there has been an increasing interest in implementing marketing techniques in the museum context, which translated into a need to have museums become more marketing oriented (Rentschler, 2002; Viglia et al., 2016; Xiang et al., 2015). Tourism practitioners and arts administrators need accurate forecasts of tourist volume, in order to effectively allocate resources and formulate pricing strategies. Our paper provides empirical evidence that a methodology based on big data has the potential to help design and implement a branding policy to address negative trends among museums across the world (Ober-Heilig, Bekmeier-Feuerhahn, & Sikkenga, 2012). Traditional research used surveys and expensive observational studies to provide data to evaluate museum visitor behavior, with limits of scale and bias. The main contribution of our research is to present a new big data approach to assess museum brand importance from the analysis of online forums, which can be correlated with data already available regarding museum visitors and their purchasing behaviors. This approach is based on the analysis of the discourse of a broad public and is less expensive than surveys. Our research offers additional evidences that can inspire researchers and practitioners in the tourism field to adopt big data methods for their decision-making processes. Museum brand managers could use metrics such as the ones included in the Semantic Brand Score to compare their brand's importance with competitor brands. They could analyze multiple sources of text data, such as social media or newspaper articles, and measure prevalence, connectivity and diversity. This study suggests to invest in marketing activities and resources that could improve online word of mouth, and increase the SBS of a brand. This means investing in a content marketing strategy that has the potential to increase the frequency with which the brand name appears within online documents (prevalence component of the SBS). Marketing managers should also carefully prepare detailed and rich content to increase the variety of information available to online users (diversity component of the SBS), as this appears to be associated with more visitors being attracted to the museum. In order to increase the connectivity of a brand within the overall conversation, managers could carefully design co-marketing strategies by partnering with institutions in the same geographic area (e.g. other museums, public sites, restaurants). The design of marketing synergies among institutions has the potential to increase the quality and depth of content offered to potential visitors, which our study shows to be associated to an increased chance of museum visits. Lastly, while it is important to monitor the sentiment of online users towards a brand, our results suggest that museum marketing managers should be less concerned about the positive or negative language used by tourists, and be more interested in improving the quality of the content provided. While most of the empirical studies thus far have been using social media and online forums to predict consumers' behaviors, the triangulation of user-generated data from various platforms represents an untapped potential. Besides looking at the total number of online comments their museum is receiving, administrators should closely review the diversity and connectivity of their brand, which our results suggest to be more impactful than sentiment.

This study extends the research on brand importance and the applications of the Semantic Brand Score that, to the extent of our knowledge, has never been used to evaluate museum brands or anticipate trends in museum visitors. Previous studies have assessed brand equity and brand importance via expensive and time consuming market surveys administered to consumers and other stakeholders, or via financial methods (Gretzel et al., 2015; Kovaleva et al., 2018). The approach we use, on the other hand, allows repeatable measurements, for a constant monitoring of brand importance with almost no additional cost. It is based on the analysis of big textual data, which come from the spontaneous conversations of tourists on online forums, without some of the biases induced by interviews (Pentland, 2010). Lastly, our findings partially contrast with studies attributing high importance to the positivity of messages for the prediction of purchasing behaviors (Kim & Ko, 2012), as sentiment in our setting was mostly uninfluential.

This study was based on a limited number of selected European museums. The sample size represents a limitation to the generalizability of the results; therefore, we recommend extending the application of this methodology to other European and non-European museums. A larger sample, where researchers could access more granular data – for example with a monthly frequency – could support the implementation of forecasting models or an in-depth time series analysis. For example, a larger dataset would allow a rolling window out-of-sample forecasting approach. Another limitation is intrinsic to all the methods based on quantitative textual analysis, since they cannot fully account for important factors impacting consumers' decisions, such as the perceived credibility of a source/reviewer or the currency of the review (i.e. how up to date is the information a reviewer is sharing). Future studies should explore the effects of other online user experience factors that might affect brand perception and economic outcome, including the official star-rating system, which is a key variable through which tourism destinations – such as museums, hotels, and restaurants – can differentiate their offering (Silva, 2015).

Twitter "I hope it is not as bad as I fear." *Procedia - Social and Behavioral Sciences*, *26*, 55–62. https://doi.org/10.1016/j.sbspro.2011.10.562